\documentclass[conference]{Conference-LaTeX-template_7-9-18/IEEEtran}
\IEEEoverridecommandlockouts
\usepackage{cite}
\usepackage{amsmath,amssymb,amsfonts}
\usepackage{graphicx}
\usepackage{textcomp}
\usepackage{xcolor}
\def\BibTeX{{\rm B\kern-.05em{\sc i\kern-.025em b}\kern-.08em
    T\kern-.1667em\lower.7ex\hbox{E}\kern-.125emX}}
\begin{document}

\title{Bridging between soft and hard thresholding by scaling
}

\author{\IEEEauthorblockN{Katsuyuki Hagiwara, \today}
\IEEEauthorblockA{\textit{Faculty of Education} \\
\textit{Mie University (Mie Univ.)}\\
1577 Kurimamachiya-cho, Tsu City, Mie Prefecture, Japan 514-8507\\
hagi@edu.mie-u.ac.jp}
}

\def\R{\mathbb{R}}
\def\E{\mathbb{E}}
\def\P{\mathbb{P}}
\def\X{{\bf X}}
\def\I{{\bf I}}
\def\0{{\bf 0}}
\def\x{{\boldsymbol{x}}}
\def\y{{\boldsymbol{y}}}
\def\i{{\boldsymbol{i}}}
\def\vbeta{{\boldsymbol{\beta}}}
\def\vvarepsilon{{\boldsymbol{\varepsilon}}}
\def\b{{\boldsymbol{b}}}
\def\evbeta{\widehat{\boldsymbol{\beta}}}
\def\ebeta{\widehat{\beta}}
\def\evb{\widehat{\b}}
\def\eb{\widehat{b}}
\def\ek{\widehat{k}}
\def\sign{{\rm sign}}
\def\ealpha{\widehat{\alpha}}
\def\eomega{\widehat{\omega}}
\def\vmu{\boldsymbol{\mu}}
\def\evmu{\widehat{\boldsymbol{\mu}}}
\def\DnST{D_n^{\rm ST}(\lambda)}
\def\DnHT{D_n^{\rm HT}(\lambda)}
\def\DnMCP{D_n^{\rm FT}(\lambda)}
\def\DnNG{D_n^{\rm NG}(\lambda)}
\def\DnNGm{D_n^{\rm SST}(\lambda)}
\def\DnNGma{D_{1,n}^{\rm SST}(\lambda)}
\def\DnNGmb{D_{2,n}^{\rm SST}(\lambda)}
\def\DnHTb{D_{2,n}^{\rm HT}(\lambda)}
\def\DnMCP{D_n^{\rm FT}}
\def\ebHT{\widehat{\beta}^{\rm HT}}
\def\ebST{\widehat{\beta}^{\rm ST}}
\def\ebMCP{\widehat{\beta}^{\rm FT}}
\def\ebNG{\widehat{\beta}^{\rm NG}}
\def\ebNGm{\widehat{\beta}^{\rm SST}}
\def\ebAL{\widehat{\beta}^{\rm AL}}
\def\eR{\widehat{R}}
\def\elambda{\widehat{\lambda}}
\def\oK{\overline{K}}
\def\eK{\widehat{K}}

\maketitle

 \begin{abstract}
Thresholding methods are simple and typical examples of sparse modeling.
Also they are successfully applied in wavelet denoising in statistical
signal processing. In this article, we developed and analyzed a
thresholding method in which soft thresholding estimators are
independently expanded by empirical scaling values.  The scaling values
have a common hyper-parameter that is an order of expansion of an ideal
scaling value that achieves hard thresholding.  We simply call this
estimator a scaled soft thresholding estimator.  The scaled soft
thresholding is a general method that includes the soft thresholding and
non-negative garrote as special cases and gives an another derivation of
adaptive LASSO.  We then derived the degree of freedom of the scaled
soft thresholding by means of the Stein's unbiased risk estimate and
found that it is decomposed into the degree of freedom of soft
thresholding and the reminder connecting to hard thresholding.  In this
meaning, the scaled soft thresholding gives a natural bridge between
soft and hard thresholding methods.  Since the degree of freedom
represents the degree of over-fitting, this result implies that there
are two sources of over-fitting in the scaled soft thresholding. The
first source originated from soft thresholding is determined by the
number of un-removed coefficients and is a natural measure of the degree
of over-fitting. We analyzed the second source in a particular case of
the scaled soft thresholding by referring a known result for hard
thresholding. We then found that, in a sparse, large sample and
non-parametric setting, the second source is largely determined by
coefficient estimates whose true values are zeros and has an influence
on over-fitting when threshold levels are around noise levels in those
coefficient estimates.  In a simple numerical example, these theoretical
implications has well explained the behavior of the degree of freedom.
Moreover, based on the results here and some known facts, we explained
the behaviors of risks of soft, hard and scaled soft thresholding
methods. These insights together with the result of a simple numerical
example showed the advantage of the bridge methods, especially the
scaled soft thresholding, in applications.
 \end{abstract}

\begin{IEEEkeywords}
bridge thresholding method, non-negative garrote, soft thresholding,
 hard thresholding, SURE
\end{IEEEkeywords}

\section{Introduction}
\renewcommand{\thefootnote}{}
\footnote[0]{\hspace{-8pt}\hrulefill\\
This paper was submitted to IEICE Trans. Inf. \& Syst.}

Regularization methods are important tools in machine
learning. Especially, there are many regularization methods for sparse
modeling such as LASSO (Least Absolute Shrinkage and Selection
Operator)\cite{LASSO}, MCP (Minimax Concave Penalty)\cite{MCP},
AL(adaptive LASSO)\cite{HZ2006}, NG (Non-negative
Garrote)\cite{LB1995,YL2007}; e.g. see also \cite{SCAD,HZ2006}. LASSO is
an $\ell_1$ penalized least squares method and has a nature of
soft-thresholding that implements thresholding and shrinkage of
coefficients; e.g. \cite{DJ1994,DJ1995}. These two properties are
simultaneously controlled by a single regularization parameter. This
causes an excessive shrinkage, thus, a large bias that is directly
related to a weakness of consistency of model selection by LASSO. This
fact has been pointed out by \cite{SCAD,LLW2006} and several methods
have been proposed for solving this problem as seen in MCP and the
others\cite{SCAD,HZ2006}. In these investigations, they discussed the
selection consistency under an appropriate choice of regularization
parameter. However, we need to choose it based on data in
applications. Cross validation is often employed to do this while there
are a few analytic approaches for model selection. For LASSO,
\cite{Dos2013,TT2012} have derived SURE (Stein's Unbiased Risk Estimate)
to choose a regularization parameter. This is possible because the
solution to LASSO is tractable while it is not explicit. Unfortunately,
in general, analytic investigation of a model selection problem is
difficult for the improvements of LASSO due to the complex and implicit
solution.

Many regularization methods for sparse modeling can be explicitly
represented as thresholding methods in case of orthogonal regression
problems. For example, it is well known that LASSO reduces to ST (Soft
Thresholding). Also, MCP and NG have simple closed forms of coefficient
estimates in orthogonal design\cite{GB1997,GAO1998,HZ2006}; e.g. FT
(Firm Thresholding) in \cite{GB1997} is a special case of MCP. Actually,
SURE for ST is well known; e.g. \cite{LARS}.  Also, SURE has been
derived for FT in \cite{GB1997} and NG thresholding in
\cite{GAO1998}. The thresholding methods are important not only in the
analysis of sparse modeling but also in applications. Especially, a
non-parametric version of orthogonal regression is important in
statistical signal processing as seen in the success of wavelet
denoising\cite{DJ1994,DJ1995}.  Due to a possibility of analytic
treatment and a significance in applications, we consider thresholding
methods of non-parametric orthogonal regression problems in this
article.

The above improvements of LASSO are intended to reduce biases of
non-zero coefficient estimates in LASSO. Therefore, the estimates move
to the least squares ones in these improvements. In this meaning, the
terminal of them is an $\ell_0$ regularization method and some of them
can be bridges for linking $\ell_1$ regularizer to $\ell_0$ one.  For
example, MCP and AL have a bridge property by controlling a
hyper-parameter while NG does not have. An $\ell_0$ regularization
method corresponds to HT (Hard Thresholding) in some sense. Therefore,
FT is a bridge that connects ST and HT. In this article, we derive a new
bridge thresholding method by introducing a scaling into ST and
investigate the degree of freedom (DOF) of the new bridge method.  Since
the degree of freedom corresponds to the degree of over-fitting (DOOF),
our investigation offers an analysis of over-fitting property of a
bridge method. Although the same investigation is possible for FT, our
method can naturally connect ST and HT, as seen later. Our attempt using
this product also reveals the properties of risk based model selection
of bridge methods.

In section II, we give a non-parametric orthogonal regression framework
including risk and SURE.  In section III, we show some existing
thresholding methods and their DOFs.  In section IV, we construct a bridge
thresholding method as a modification of ST and derive its DOF. In
section V, we show numerical examples for investigating the DOF for
which we explain the behavior of the DOF based on our result in section
IV.  Section VI is devoted for conclusions and future works.

\section{Setting}

\subsection{Non-parametric orthogonal regression}

Let $(\x_i,y_i),~i=1,\ldots,n$ be samples in which
$\x_i=(x_{i,1},\ldots,x_{i,n})$.  Let $\X$ be an $n\times n$ design
matrix whose $(i,j)$ element is $x_{i,j}$. We define
$\y=(y_1,\ldots,y_n)^T$, where $~^T$ denotes the transpose of a
matrix. We here consider a regression problem by $\X\vbeta$ of $\y$,
where $\vbeta=(\beta_1,\ldots,\beta_n)^T$ is a coefficient vector. This
is a non-parametric regression problem. We especially assume that the
orthogonality of design; i.e. $\X^T\X=n\I_n$, where $\I_n$ is an
$n\times n$ identity matrix. A typical example of this problem is
discrete wavelet transform in signal processing.  For example, in
time series applications, we set $x_{i,j}=g_j(t_i)$, where $g_j$ is a
univariate wavelet function and $t_i$ is the $i$th sampling time.

We assume that $\y$ is generated by the rule : 
\begin{equation}
\label{eq:y} 
\y=\X\b+\vvarepsilon,
\end{equation}
where $\b=(b_1,\ldots,b_n)^T$ and
$\vvarepsilon=(\varepsilon_1,\ldots,\varepsilon_n)^T$.  We also assume
i.i.d. Gaussian additive noise; i.e.  $\vvarepsilon\sim
N(\0_n,\sigma^2\I_n)$, where $\0_n$ is an $n$-dimensional zero vector
and $\sigma^2$ is a noise variance.  Here, $\b$ is a true coefficient
vector.  We define $K^*=\{k:b_k\neq 0\}$ and $k^*=|K^*|$ that is the
number of members of $K^*$.  In a sparse setting, we may assume that
$k^*\ll n$.  It is easy to see that
\begin{equation}
\evb=\frac{1}{n}\X^T\y
\end{equation}
is the least squares estimate and $\y=\X\evb$; i.e. it is a
transformation of $\y$. We have $\evb\sim N(\b,\sigma^2\I_n/n)$ under
the assumption on $\vvarepsilon$.  Therefore, $\eb_1,\ldots,\eb_n$ are
mutually independent and $\eb_k\sim N(b_k,\tau_n^2)$, where we define
$\tau_n=\sigma/\sqrt{n}$.

\subsection{Risk and degree of freedom}

Let $\evbeta=(\ebeta_1,\ldots,\ebeta_n)^T$ be an estimate that is
calculated by $\evb$; i.e. $\evbeta=\evbeta(\evb)$ that is a
modification of the least squares estimate.
We define $\evmu=\X\evbeta$ and $\vmu=\X\b$.
Let $\E_{\y}$ denotes the expectation with respect to the joint
probability distribution of $\y$.
We define the risk by
 \begin{align}
  R_{n}&=\frac{1}{n}\E_{\y}\|\evmu-\vmu\|^2
 \end{align}
By defining
\begin{align}
D_{n}=\E_{\y}(\evmu-\E_{\y}\evmu)^T(\y-\vmu),  
\end{align}
it is easy to see that
 \begin{align}
  R_{n}&=\frac{1}{n}\E_{\y}\|\evmu-\y\|^2-\sigma^2
  +\frac{2}{n}D_n
 \end{align}
holds, where we used $\E_{\y}\y=\vmu$ and $\y\sim N(\vmu,\sigma^2\I_n)$.
$D_{n}$ is the covariance between $\evmu$ and $\y$.  Since
$\vvarepsilon=\y-\vmu$, the DOF can be regarded as a measure of
over-fitting (to noise). $D_{n}$ is called the degree of freedom (DOF).
It is also called the optimism in \cite{RT2014}.  From the above
viewpoint, it can be also regarded as the degree of over-fitting (DOOF).
If the model complexity increases then DOOF is high and, thus, the DOF
is high. As a typical example, the number of variables is a measure of
the model complexity. Actually, as well known in case of the least
squares estimation, the DOF is the sum of the variances of estimators
and in proportion to the number of variables.

In the expectation, we can replace $\y$ with $\evb$ by the change of
variables and further have
\begin{align}
 D_{n}&=n\E_{\evb}(\evbeta-\E\evbeta)^T(\evb-\b)\notag\\
 &=n\E_{\evb}\evbeta^T(\evb-\b)
\end{align}
since $\E_{\y}\evb=\b$, where $\E_{\evb}$ is the expectation with respect to
the joint distribution of $\evb$. We further assume that $\ebeta_k$ is
calculated only by $\eb_k$; i.e.  $\ebeta_k=\ebeta_k(\eb_k)$ and the
others are not included in calculating $\ebeta_k$.  The estimates
considered in this article satisfy this condition. In this case, we can
write
\begin{align}
 D_{n}&=n\sum_{k=1}^n\E_{\eb_k}\ebeta_k(\eb_k-b_k),
\end{align}
where $\E_{\eb_k}$ is the expectation with respect to the marginal
distribution of $\eb_k$.

In case of the least squares method, as well known, the DOF corresponds
to the variance of the estimator; i.e. in case of
$\ebeta_k(\eb_k)=\eb_k$. However, this does not hold in general.

Since $\eb_k\sim N(b_k,\sigma^2/n)$, the Stein's lemma for this case
states that
 \begin{align}
\label{eq:dof-stein-1}  
\E_{\eb_k}\ebeta_k(\eb_k-b_k)
 =\E_{\eb_k}\frac{\partial \ebeta_k}{\partial\eb_k}
 \end{align}
 holds if $\ebeta_k=\ebeta_k(\eb_k)$ is absolutely continuous as a
function of $\eb_k$ and, of course, the right hand side
exists. Therefore, we have
 \begin{align}
\label{eq:dof-stein-2}   
D_{n}&=\sigma^2
  \sum_{k=1}^n\E_{\eb_k}\frac{\partial \ebeta_k}{\partial\eb_k}
=\sigma^2\E_{\evb}
 \sum_{k=1}^n\frac{\partial \ebeta_k}{\partial\eb_k}.  
 \end{align}
This tells us that the DOF is determined by the degree of variation of
$\ebeta_k$ with respect to $\eb_k$. This is natural since the DOF is the
covariance between a modified estimator and the least squares estimator;
i.e. $\evbeta(\evb)$ and $\evb$. The covariance can be large when the
variation of $\ebeta_k$ is large. For example, if we simply consider
$\ebeta_k=\gamma\eb_k$ with $\gamma>1$ then the DOF is proportional to
$\gamma$ while such an estimation cannot show a good performance.  If
the degree of variation changes with $\eb_k$ then the DOF is large when
a large variation occurs around the point at which the probability
density is high.

The unbiased estimate of $R_n$ with the representation of
(\ref{eq:dof-stein-2}) is called Stein's Unbiased Risk Estimate (SURE).
We simply use $\E$ for $\E_{\evb}$ and $\E_k$ for $\E_{\eb_k}$ below.

\section{Some existing thresholding methods and their SUREs}

We here list some existing thresholding methods that are induced by
regularization method. Those include HT, ST, NG and FT. All of these
are explicitly represented as a function of $\eb_k$ for each $k$. 
For a set $E$, we denote the complement of $E$ by $\overline{E}$.
We define
\begin{equation}
I_{E}(u)=\begin{cases}
                1 & u\in E\\
                0 & u\notin E\\
               \end{cases}.
\end{equation}
for $u\in\R$ and $E\subseteq\R$.
We define
\begin{align}
I_{\lambda}(u)=1-I_{(-\lambda,\lambda)}(u),
\end{align}
which is an index function on $\overline{(-\lambda,\lambda)}$.
We also define $(u,0)_{+}=\max(u,0)$.

\subsection{Hard thresholding (HT)}

We define a hard thresholding function by
\begin{equation}
H_{\lambda}(u)=uI_{\lambda}(u),
\end{equation}
where $\lambda>0$ is a parameter that is a threshold level.
The HT estimator is obtained by
\begin{equation}
\ebHT_{k}=H_{\lambda}(\eb_k).
\end{equation}
This corresponds to an $\ell_0$ regularized estimate from a view point of
regularization. We define
\begin{equation}
\label{eq:active-set-HT} 
\eK_{\lambda}=\{k:|\eb_k|\ge\lambda\}
\end{equation}
and $\ek_{\lambda}=|\eK_{\lambda}|$. Note that $\eK_{\lambda}$ is often
called an active set or a support.  Since $H_{\lambda}$ is not
continuous, we cannot apply the Stein's lemma. As shown in \cite{RT2014}
and also in the later section, it is possible to evaluate the DOF of HT
by using different ways. However, we find that it is useless in
applications since we need information on a true representation for
constructing SURE. Instead of applying SURE,
there are some proposals of determining an optimal threshold level of
HT; e.g. \cite{DJ1994,DJ1995}.

\subsection{Soft thresholding (ST)}

Let $\sign$ be a sign function.
We define a soft thresholding function by
\begin{equation}
 S_{\lambda}(u)=
\begin{cases}
 u-\sign(u)\lambda & |u|\ge\lambda\\
 0 & |u|<\lambda,
\end{cases}
\end{equation}
where $\lambda>0$ is a parameter for simultaneously determining both of
threshold level and amount of shrinkage. $S_{\lambda}$ can be written
as
\begin{align}
 S_{\lambda}(u)&=\left(1-\lambda/|u|\right)_{+}u.
\end{align}
Note that, here, $1-\lambda/|u|$ is an amount of shrinkage for
$u\ge\lambda$.

We employ
\begin{equation}
 \ebST_k=S_{\lambda}(\eb_k)
\end{equation}
This is an $\ell_1$ regularized estimate from a view point of regularization.
Note that the active set of ST is consistent with that of HT for the
same value of $\lambda$.
The estimates in the active set are shrunk towards zero in ST by $\lambda$.
We denote the DOF of ST by $\DnST$.
Since we have
\begin{equation}
 \frac{\partial\ebST_k}{\partial\eb_k}=I_{\lambda}(\eb_k),
\end{equation}
we obtain
\begin{align}
\label{eq:dof-st}
 \DnST&=
\sigma^2\E
 \sum_{k=1}^nI_{\lambda}(\eb_k)=\sigma^2\E\ek_{\lambda}
\end{align}
by the Stein's lemma. This result is typically useful in applications
since we have SURE :
\begin{align}
 \eR_n=\frac{1}{n}\|\y-\X\evbeta^{\rm ST}\|^2-\sigma^2
 +\frac{2\sigma^2\ek_{\lambda}}{n},
\end{align}
where $\evbeta^{\rm ST}=(\ebST_1,\ldots,\ebST_n)^T$.  This can be $C_p$-type
model selection criterion\cite{Cp} when we can obtain an appropriate
estimate of $\sigma^2$ in applications; e.g. median absolute deviation
(MAD) estimate for the first detail coefficients in
wavelet decomposition\cite{DJ1994,DJ1995}.

\subsection{Nnon-negative garrote (NG)}

NG is a regularization method\cite{LB1995,YL2007,HZ2006} and can be
explicitly solved in an orthogonal case\cite{LB1995,GAO1998}, which has
been given by
\begin{align}
 \ebNG_k&=(1-\lambda^2/|\eb_k|^2)_{+}\eb_k.
\end{align}
It is easily understood that NG can relax
a bias problem of ST at large absolute value of coefficient
estimates. If we define
\begin{equation}
 \ealpha_k=
\begin{cases}
1+\lambda/|\eb_k| & |\eb_k|\ge\lambda\\
2 & |\eb_k|<\lambda
\end{cases}
\end{equation}
then it is easy to see that the NG estimate is written as
\begin{equation}
\ebNG_k=\ealpha_kS_{\lambda}(\eb_k).
\end{equation}
In the definition of $\ealpha_k$, the value for $|\eb_k|<\lambda$ is not
important since $S_{\lambda}(\eb_k)=0$ in this area. The definition
guarantees the continuity of $\ealpha_k$ formally.
This is a scaled ST estimate with a scaling value $\ealpha_k$\cite{KH2017}.
We denote the DOF for NG by $\DnNG$.
Since we have
\begin{align}
 \frac{\partial\ebNG_k}{\partial\eb_k}&=
(1+\lambda^2/|\eb_k|^2)I_{\lambda}(\eb_k),
\end{align}
we obtain
\begin{align}
 \DnNG&=\sigma^2\E\ek_{\lambda}
 +\sigma^2\E\sum_{k\in\eK_{\lambda}}
\frac{\lambda^2}{\eb_k^2}
\end{align}
by the Stein's lemma; e.g. see also \cite{GAO1998}.
Note that the first term of $\DnNG$ is the DOF of ST.

\subsection{Firm thresholding (FT) }

MCP\cite{MCP} is a regularization method with non-convex penalty, by
which a bias problem of LASSO is known to be relaxed. In the orthogonal
case, MCP reduces to FT\cite{GB1997}.

We define a function by
\begin{equation}
F_{\gamma,\lambda}(u)=\begin{cases}
                u & |u|\ge\gamma\lambda\\
                \frac{\gamma}{\gamma-1}S_{\lambda}(u)
                       & |u|<\gamma\lambda\\
               \end{cases},
\end{equation}
where $\lambda>0$ and $\gamma>1$ are parameters.
Note that $\gamma\lambda\ge\lambda$ since $\gamma>1$.
The FT estimate in an orthogonal case is given by
\begin{equation}
 \ebMCP_k(\eb_k)=F_{\gamma,\lambda}(\eb_k).
\end{equation}
By the definition of $F_{\gamma,\lambda}$, it is easily understood that
the estimates with large absolute values are harmless in FT.

We define
\begin{align}
\eK_{0,\gamma,\lambda}&=\{k:|\eb_k|\ge\gamma\lambda\}\\
\eK_{1,\gamma,\lambda}&=\{k:\lambda\le|\eb_k|<\gamma\lambda\}
\end{align}
and $\ek_{j,\gamma,\lambda}=|\eK_{j,\gamma,\lambda}|$ for $j=0,1$.
Since we have
\begin{equation}
 \frac{\partial\ebeta_k}{\partial\eb_k}=
  \begin{cases}
   1 & |u|\ge\gamma\lambda\\
   \frac{\gamma}{\gamma-1} & \lambda\le|u|\le\gamma\lambda\\
   0 & |u|<\lambda\\
  \end{cases},
\end{equation}
the DOF of FT is given by
\begin{align}
 \DnMCP(\gamma)&=\sigma^2
 \E\sum_{k=1}^nI_{\gamma\lambda}(\eb_k)+\frac{\gamma}{\gamma-1}
 \E\sum_{k=1}^nI_{(\lambda,\gamma\lambda]}(|\eb_k|)\notag\\
 &=\sigma^2\E\ek_{0,\gamma\lambda}+
 \sigma^2\frac{\gamma}{\gamma-1}\E\ek_{1,\gamma,\lambda}
\end{align}
by the Stein's lemma.

Note that FT connects ST to HT by controlling $\gamma$; i.e. FT goes to
ST as $\gamma\to\infty$ and goes to HT as $\gamma\to 1$ in some
sense. On the other hand, NG does not have this property.  However, as
shown in the next section, we can extend NG to bridge the gap between ST
and HT. It is a natural extension of ST and can
connect to HT by a hyper-parameter.

\subsection{Adaptive LASSO (AL)}

In \cite{HZ2006}, adaptive LASSO estimate in an orthogonal case 
has been given by
\begin{align}
 \ebAL(\gamma)=\left(|\eb_k|-\lambda_R/|\eb_k|^{\gamma}\right)_{+}
 \sign(\eb_k)
\end{align}
where $\gamma>0$ and $\lambda_R>0$ is a regularization parameter.
This
can be written as
\begin{align}
\ebAL(\gamma)=\left(1-\lambda_R/|\eb_k|^{\gamma+1}\right)_{+}\eb_k.
\end{align}
The threshold level is, thus, given by $\lambda=\lambda_R^{1/(\gamma+1)}$
in AL.  Therefore, AL is essentially equivalent to NG when $\gamma=1$
if we choose $\lambda_R=\lambda^{2}$ under a given $\lambda$.
We show SURE for AL in later section.

\section{Scaling of soft thresholding estimator}

\subsection{Scaling of soft thresholding estimator}

We focus on a scaling of the ST estimate :
\begin{equation}
\alpha_kS_{\lambda}(\eb_k),
\end{equation}
where $\alpha_k\ge 0$. We define
\begin{align}
\label{eq:hd-scaling}
\eomega_k=\frac{1}{1-\lambda/|\eb_k|}
\end{align}
for $|\eb_k|>\lambda$.
It is easy to check that 
$\eomega_kS_{\lambda}(\eb_k)=\eb_k$ holds for
$|\eb_k|>\lambda$. This implies that $\eomega_k$ is a scaling value that
gives us a hard thresholding estimate when we give an appropriate definition
at $\lambda$. If
$|\eb_k|>\lambda$ then we have
\begin{align}
\eomega_{k}=1+(\lambda/|\eb_k|)+(\lambda/|\eb_k|)^2+(\lambda/|\eb_k|)^3+\cdots
\end{align}
by the Taylor expansion. This implies that the NG estimate is obtained
by the first order approximation of $\eomega_k$ when we regard the NG
estimate as a scaling of the ST estimate.  We consider the $m$th order
approximation of $\eomega_k$ and employ
\begin{align}
 \ealpha_{k,m}=
\begin{cases}
1+\sum_{j=1}^m
 \lambda^j/|\eb_k|^j & |\eb_k|\ge\lambda\\
m+1 & |\eb_k|<\lambda
\end{cases}
\end{align}
as a scaling value. This is well defined for $\eb_k\ge\lambda$ if $m<\infty$.
If $|\eb_k|\ge\lambda$ then it is easy to check that
\begin{align}
 \ealpha_{k,m}S_k(\eb_k)=\eb_k-\frac{\lambda}{\eb_k}
 \frac{\lambda^m}{|\eb_k|^{m-1}}
\end{align}
holds for $m\ge 1$. If $m$ is odd then we simply have
 \begin{align}
\label{eq:SST-0}  
 \ealpha_{k,m}S_k(\eb_k)=\eb_k-\frac{\lambda^{m+1}}{\eb_k^{m}}
 \end{align}
if $|\eb_k|\ge\lambda$.
We thus have an estimate by
 \begin{align}
\label{eq:SST}    
 \ebNGm_k=\ealpha_{k,m}S_k(\eb_k)
 =\left(1-\lambda^{m+1}/\eb_k^{m+1}\right)_{+}\eb_k
 \end{align}
when $m$ is odd.  We refer to this method as a scaled soft thresholding
 estimator; i.e. SST estimator for short.  Obviously, the SST estimator
 expands the ST estimator while it shrinks the least squares estimator.
 As was seen above, SST is consistent with AL if we choose
 $\lambda_R=\lambda^{\gamma+1}$ in AL.  Therefore, SST is an another
 implication of AL, in which a hyper parameter in AL corresponds to the
 order of expansion of scaling values of ST. We denote the DOF for SST
 by $\DnNGm$.  Since we have
\begin{equation}
 \frac{\partial\ebNGm_k}{\partial\eb_k}=
\begin{cases}
 1+m\lambda^{m+1}/\eb_k^{m+1} & |\eb_k|\ge\lambda\\
0 & |\eb_k|<\lambda
\end{cases},
\end{equation}
we obtain
\begin{align}
 \DnNGm&=\DnNGma+\DnNGmb
\end{align}
by the Stein's lemma, where
 \begin{align}
  \label{eq:deng-1}
  \DnNGma&=\sigma^2\sum_{k=1}^n\E_kI_{\lambda}(\eb_k)
  =\sigma^2\E\ek_{\lambda}\\
  \label{eq:deng-2}
  \DnNGmb&=\sigma^2\sum_{k=1}^nm\E_k
  \left(\frac{\lambda}{\eb_k}\right)^{m+1}I_{\lambda}(\eb_k)\notag\\
  &=\sigma^2m\E\sum_{k\in\eK_{\lambda}}
\left(\frac{\lambda}{\eb_k}\right)^{m+1}
 \end{align}
 when $m$ is odd. In (\ref{eq:deng-2}), we define that the sum is zero
   if $\eK_{\lambda}$ is empty. Note that if we set
   $\lambda=\lambda_R^{1/(\gamma+1)}$ then $\DnNGm$ is the DOF of AL.

\begin{figure}[t]
 \begin{center}
\includegraphics[width=65mm]{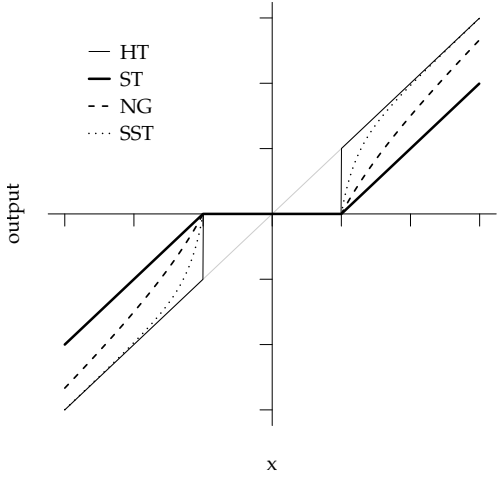}
\caption{Outputs of HT, ST, NG and SST with $m=5$, where
  $\lambda=1.0$.}
\label{fig:thfunc2}
 \end{center}
\end{figure}

\subsection{Discussion on SST}

SST includes ST if $m\ge 1$.  SST reduces to formally ST if $m=0$ and it
reduces to a naive NG if $m=1$. We show the output of HT, ST, NG and SST
with $m=5$ in Fig.\ref{fig:thfunc2}. It is obvious that, by controlling
$m$, SST can be bridge for linking ST and HT in some sense. Note that
the first term of $\DnNGm$ is the DOF of ST. Therefore, the second term
is related to HT.  Since the second term is positive, the DOF of SST is
larger than that of ST.  Since the DOF represents the DOOF, this result
says that SST has two sources of over-fitting and the over-fitting is
more serious in SST compared to ST. The first source that is originated
from the first term is the DOF of ST and, thus, is determined by the
number of non-zero coefficient estimates as seen in
(\ref{eq:dof-st}). This is a natural measure of the model complexity,
thus the DOOF. The second one that is originated from
the second term is related to the property of HT; i.e. other than the
property of ST. This does not seem to be a simple form that reflects the
model complexity. $(\lambda/\eb_k)^{m+1}$ is large when $\eb_k$ is close
to $\lambda$. This is obvious from the fact that the variation of the
second term of (\ref{eq:SST-0}) with respect to $\eb_k$ is large when
$\eb_k$ is close to $\lambda$ with $\lambda\ge |\eb_k|$. Since $\eb_k$ is
locally distributed around $b_k=\E\eb_k$, the contribution of the $k$th
component is large if $\lambda$ is close to $\E\eb_k$. This corresponds
to the implication of (\ref{eq:dof-stein-1}) derived by the Stein's
lemma. In other words, if we change $\lambda$ then the impact of the
$k$th component is large in $\DnNGmb$ when $\lambda$ is around $b_k$.
Especially, there are many components with $b_k=0$ in a sparse
setting. As a result, it is expected that $\DnNGmb$ takes a large value
when $\lambda$ is close to zero while it may not be straightforward. We
consider the behavior of $\DnNGmb$ in detail through the analysis of HT
since HT may approximate SST with a large $m$ in some sense.

\subsection{Limit of DOF of SST}

Since SST approaches HT as $m\to\infty$ in some sense, we firstly
consider the DOF of HT. Then, we discuss the convergence of the DOF of
SST to that of HT.  The DOF of HT has already been given in
\cite{RT2014}. We here show a simple derivation of the DOF of HT.
\begin{align}
\DnHT&=n\sum_{k=1}^n\E_{k}\ebHT(\eb_k-b_k).
\end{align}
By defining
\begin{align}
M_{\lambda}(u)=\lambda\left\{
 I_{[\lambda,\infty)}(u)-I_{(-\infty,-\lambda]}(u)\right\},
\end{align}
we have
 \begin{align}
  \label{eq:H-S-relation}
 H_{\lambda}(u)=S_{\lambda}(u)+M_{\lambda}(u).
 \end{align}
We then have
 \begin{align}
  \label{eq:DnHT}
  \DnHT&=\DnST+\DnHTb,
 \end{align}
where 
\begin{align}
\label{eq:DnHTb}
 \DnHTb&=n\sum_{k=1}^n\E_{k}(\eb_k-b_k)M_{\lambda}(\eb_k).
\end{align}
This is called the ``search degrees of freedom'' of best subset
selection in \cite{RT2014}. This may because $M_{\lambda}$ seems to
behave as an index function for choosing non-zero components;
i.e. subset search. Let $\phi_{\mu,\tau}$ be a
probability density function of $N(\mu,\tau^2)$. It is easy to see that
\begin{align}
\label{eq:DnHTb-integ-1}
 \int_{\lambda}^{\infty}(\xi-\mu)\phi_{\mu,\tau}(\xi)d\xi
 &=\tau^2\phi_{\mu,\tau}(\lambda)\\
\label{eq:DnHTb-integ-2}
 \int_{-\infty}^{-\lambda}(\xi-\mu)\phi_{\mu,\tau}(\xi)d\xi
 &=-\tau^2\phi_{\mu,\tau}(-\lambda)
\end{align}
hold. We define 
 \begin{align}
\label{eq:h_bk_taun}  
h_{\mu,\tau}(\lambda)= 
\lambda\left(\phi_{\mu,\tau}(\lambda)
 +\phi_{\mu,\tau}(-\lambda)\right).
 \end{align}
Since $\eb_k\sim N(b_k,\tau_n^2)$ and $\tau_n^2=\sigma^2/n$, we then have
  \begin{align}
   \label{eq:DnHTb-result}
   \DnHTb&=\sigma^2
  \sum_{k=1}^nh_{b_k,\tau_n}(\lambda)
  \end{align}
by (\ref{eq:DnHTb}), (\ref{eq:DnHTb-integ-1}) and
(\ref{eq:DnHTb-integ-2}).  As an another expression, by using the
Dirac's delta function, we can write
\begin{align}
\label{eq:DnNGmb-delta} 
 \DnHTb=
\sigma^2\lambda\sum_{k=1}^n
 \E_k\left(\delta(\eb_k-\lambda)+\delta(\eb_k+\lambda)\right).
\end{align}
The extension of the Stein's lemma in this direction have been made in
\cite{RT2014}. This gives us a natural interpretation in the sense of
the derivative of the step function in (\ref{eq:H-S-relation}).
Unfortunately, this result implies that a risk estimate for HT is not
applicable since it needs information on a true representation. For this
reason, \cite{DPF2013} has considered an approximation of the DOF of HT
based on a relaxation using a Gaussian kernel. The important point here
is that the DOF of HT is also decomposed into the DOF of ST and the
extra term.

It is easy to show that the DOF of HT is obtained as a limit of that of SST.
To do this, we just evaluate $\DnNGmb$ in SST since the first term that
is the DOF of ST is common for both. We define
$g_{m.\lambda}(x)=m\lambda^{m}/x^{m+1}$.  We assume that $m$ is odd and
$m>1$.  We then have
\begin{align}
 \label{eq:integral-of-gm}
\int_{\lambda}^{\infty}g_{m,\lambda}(x)=1.
 \end{align}
By the mean value theorem, there exits $\theta\in(0,1)$ such that
 \begin{align}
 \label{eq:mvtheory-for-phi}  
 \phi_{\mu,\tau}(t)-\phi_{\mu,\tau}(\lambda)=
 (t-\lambda)\phi_{\mu,\tau}'((1-\theta)\lambda+\theta t)
 \end{align}
for $t\ge\lambda$, where $'$ stands for a derivative. It is obvious that
there exists a positive constant $K$ that satisfies
$|\phi_{\mu,\tau}'(t)|<K$ for any $t\in\R$.  By
(\ref{eq:integral-of-gm}) and (\ref{eq:mvtheory-for-phi}),
we have
\begin{align}
& \left|\int_{\lambda}^{\infty}g_{m,\lambda}(t)\phi_{\mu,\tau}(t)dt
 -\phi_{\mu,\tau}(\lambda)
 \right|\notag\\
 =& \left|\int_{\lambda}^{\infty}g_{m,\lambda}(t)\left\{
\phi_{\mu,\tau}(t)-\phi_{\mu,\tau}(\lambda)
\right\}dt
 \right|\notag\\
 \le&
 K\left|\int_{\lambda}^{\infty}(t-\lambda)g_{m,\lambda}(t) dt \right|
 =K\frac{\lambda}{m-1}.
\end{align}
Therefore, we have
\begin{align}
 \lim_{m\to\infty}
 \int_{\lambda}^{\infty}g_{m,\lambda}(t)\phi_{\mu,\tau}(t)dt=
 \phi_{\mu,\tau}(\lambda).
\end{align}
Since $m$ is odd, the same argument yields
\begin{align}
 \lim_{m\to\infty}
 \int_{-\infty}^{-\lambda}g_{m,\lambda}(t)\phi_{\mu,\tau}(t)dt
 =\phi_{\mu,\tau}(-\lambda).
 \end{align}
We thus have
\begin{align}
 \DnNGmb
 &=\sigma^2\sum_{k=1}^n\E_k
\lambda g_{m}(\eb_k)I_{\lambda}(\eb_k)\notag\\
 &\to \sigma^2\sum_{k=1}^nh_{b_k,\tau_n}(\lambda)~~~(m\to\infty),
\end{align}
for $\lambda>0$. Therefore, we can obtain
\begin{align}
 \lim_{m\to\infty}\DnNGm=\DnST+\DnHTb=\DnHT.
\end{align}
Note that the above derivation can be applied to the other bridge
 methods including FT. Actually, the above result is also obtained as a
 limit of $\gamma\to 1$ in FT. However, SST has an advantage in this
 analysis since it gives us a natural decomposition of the DOF; i.e.
 the DOF of ST and the reminder connecting to HT. This is because SST
 includes ST as a part of the Taylor expansion. Moreover, the first term
 that is the DOF of ST is consistent with that of HT.  Such an exact and
 convenient decomposition of the DOF does not appeared in FT. By this
 result, we can guarantee that the DOF of SST with a large $m$ is
 approximated by that of HT.
 
\section{A numerical example and discussions}

\subsection{Numerical example}

We consider a set of $n$ functions,
$G_n=\left\{g_1,g_2,\ldots,g_n\right\}$, in which
\begin{align}
g_k(t)&=
\begin{cases}
1 & k=1\\
\sqrt{2}\cos(k t/2) & \mbox{$k$ : even and $k\neq 1,n$}\\
\sqrt{2}\sin(k t/2).& \mbox{$k$ : odd and $k\neq 1,n$}\\
\cos(k t/2) & k=n\\
\end{cases}.
\end{align}
We set $t_i=2\pi (i-1)/n$ for $i=1,\ldots,n$, where $n$ is even.  We
then choose $x_{i,j}=g_j(t_i)$.  Then, $\X$ is an $n\times n$ orthogonal
matrix.

We here conduct a monte carlo simulation.  The result here is partly
consistent with the experiment in \cite{RT2014}; i.e. the result for HT.
We generate samples according to (\ref{eq:y}) under the condition below.
We employ the above design matrix. We set $K^*=\{1,2,3,4,5\}$ and
$b_k=1$, $k=1,2,3,4,5$ for a true representation. We set $\sigma^2=1$
for a Gaussian additive noise.  We set $n=256$. We then estimate the
coefficients by using HT, ST and SST with $m=21$ under a fixed threshold
level. Here, we refer to SST with $m=21$ as SST simply.  Note that
$m=21$ may be large enough for approximating HT.

For the coefficient estimates, we calculate SURE and an actual risk. The
latter is calculated by the squared error sum between the estimated
outputs and true outputs. We repeat this procedure for $S=5000$ times
and calculate the averages of SUREs and actual risks at a fixed
threshold level.  We conduct this procedure for candidates of the
threshold level,
$\lambda\in\{0.01,0.02,\dots,0.1,0.15,0.2,\ldots,1,2,\ldots 10\}$.

\begin{figure}[t]
 \begin{center}
  \includegraphics[width=75mm]{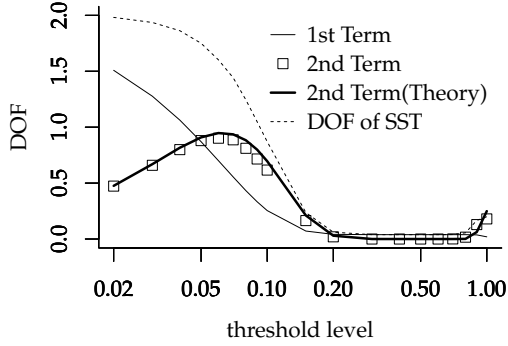}

  (a) DOF of SST
  
  \includegraphics[width=75mm]{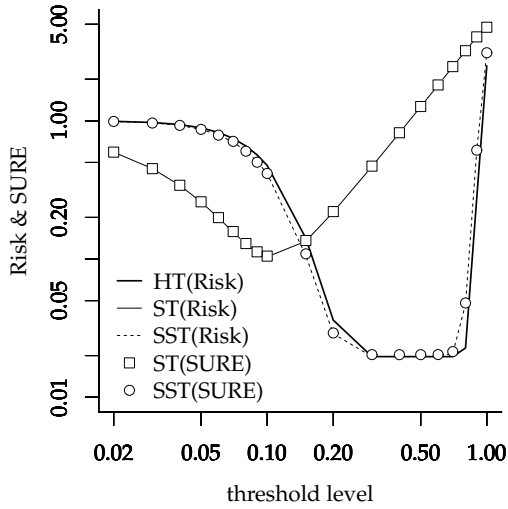}

  (b) Risk and SURE

  \caption{The results of monte carlo simulation.  (a) The first and
second terms of the DOF of SST, the entire DOF of SST and the
theoretical value of the second term for HT. We set $m=21$ for SST in
this simulation. (b) Risks for HT, ST and SST together with SUREs for
ST and SST. } 
\label{fig:verification}
 \end{center}
\end{figure}

The results are summarized in Fig.\ref{fig:verification}.
Fig.\ref{fig:verification} (a) shows the first and second terms of the
DOF of SST, the entire DOF of SST and the theoretical value of the
second term for HT.  Fig.\ref{fig:verification} (b) shows the risks for
HT, ST and SST together with SUREs for ST and SST.
We explain the validity of this result below.

\subsection{Discussion on DOF}

In Fig.\ref{fig:verification} (a), we can see that the first term of DOF
decreases as the threshold level, $\lambda$, increases. This is because
it is the DOF of ST as (\ref{eq:dof-st}) says.  On the other hand, the
second term of the DOF of SST is almost consistent with the theoretical
value for HT.  This supports our theoretical result for a large $m$ in
SST above. The important point in Fig.\ref{fig:verification} (a) is 
the non-monotonicity of the second term of DOF for SST. It is a
different property seen in ST and is mentioned in \cite{RT2014}.
We next consider this point for HT, thus SST with a large $m$.

Here, we remind the definition of $K^*=\{k:b_k\neq 0\}$ and $k^*=|K^*|$.
We consider a sparse and large sample case, in which $k^*\ll n$ and $n$
is large. In this setting, $\tau_n^2=\sigma^2/n$ that is the variance of
$\eb_k$ is small since $n$ is large. By (\ref{eq:h_bk_taun}), it is
easy to see that $h_{0,\tau_n}(\lambda)\ge 0$ and
\begin{align}
 h_{0,\tau_n}'(\lambda)=&2\left(1-\frac{\lambda^2}{\tau_n^2}\right)
 \phi_{b_k,\tau_n}(\lambda)
\end{align}
holds; e.g. see also \cite{RT2014}.  Therefore, for components with zero
true coefficients, $h_{b_k,\tau_n}$ increases for $\lambda<\tau_n$ and
decreases for $\lambda>\tau_n$ as $\lambda$ increases; i.e. the peak
occurs at $\lambda=\tau_n$. This non-monotonicity of the DOF is a
different property seen in ST and is mainly mentioned in \cite{RT2014}.
We now step into the analysis of the DOF of HT under a sparse, large
sample and non-parametric setting. We consider the case where
$\lambda=\tau_n$. If we define $S_1(\lambda)=\sum_{k\in
K^*}h_{b_k,\tau_n}(\lambda)$ and $S_2(\lambda)=\sum_{k\in
\oK^*}h_{b_k,\tau_n}(\lambda)$ then $\DnHTb=\sigma^2(S_1+S_2)$. $S_1$
and $S_2$ are due to the non-zero true components and zero true
components respectively. We roughly evaluate $S_1(\tau_n)$ and
$S_2(\tau_n)$ in a sparse and large sample setting. If $k\in K^*$ then
we have $\tau_n(\phi_{b_k,\tau_n}(\tau_n)+\phi_{b_k,\tau_n}(-\tau_n))
\simeq C_1e^{-nC_2}$ for positive constants $C_1$ and $C_2$ since
$\tau_n=\sigma/\sqrt{n}\ll |b_k|$. We thus have $S_1(\tau_n)\simeq
k^*C_1e^{-nC_2}$.  On the other hand, if $k\in \oK^*$ then
$\tau_n\phi_{b_k,\tau_n}(\tau_n)=C_3$ for a positive constant $C_3$. We
thus have $S_2(\tau_n)=2(n-k^*)C_3\simeq 2nC_3$ in a sparse
setting. Therefore, we have $S_1(\lambda)\ll S_2(\lambda)$ at
$\lambda=\tau_n$; i.e. the impact of zero true components dominates
those of non-zero true components around $\lambda=\tau_n$.  The impact
of the zero true coefficients monotonically decreases as $\lambda$
increases for $\lambda\ge \tau_n$. For example, if we define
$\lambda_n=\sqrt{2\sigma ^2\log n/n}$ and set $\lambda=\lambda_n$ then
$\tau_n\phi_{0,\tau_n}(\tau_n)\simeq C/n$ for a positive constant
$C$. Therefore, we have $S_2(\lambda_n)\simeq 2C$ which is very smaller
than $S_2(\tau_n)$. For the non-zero true components, if $\lambda$ is
around $|b_k|\neq 0$ then either $\phi_{b_k,\tau_n}(\lambda)$ or
$\phi_{b_k,\tau_n}(-\lambda)$ is large. Therefore, $S_1(\lambda)$ can be
large around $\lambda=|b_k|$. If $\lambda=|b_k|$ then
$h_{b_k,\tau_n}(\lambda)\simeq C\sqrt{n}$ for a positive constant
$C$. However, if $\lambda=|b_k|\pm\epsilon$ for $\epsilon>0$ then
$h_{b_k,\tau_n}(\lambda)\simeq \sqrt{n}C_1e^{-nC_2}$ for positive
constants $C_1$ and $C_2$.  Therefore, the impact of a non-zero true
coefficient is point-wise.  Additionally, it is a single contribution at
only around $\lambda=|b_k|\neq 0$; i.e. in general, $b_k$, $k\in K^*$
take different values. This is different from a mass and large effect of
zero true components at around $\lambda=\tau_n$.

As a result, we can say that the effect of the second source of
over-fitting in HT and, thus SST with a large $m$ can be large around
$\lambda=\tau_n$ in a sparse and large sample setting. 
And it is brought about by estimates of zero true components.
In our setting of
the numerical example, we have $\tau_n=\sigma/\sqrt{n}=0.0625$ and the
second term is maximized at around this value in
Fig.\ref{fig:verification} (a). This is an evidence for our discussion
above.  Note that $\tau_n=\sigma/\sqrt{n}$ is small in a large sample
case. Therefore, the second term is relatively large for small value of
$\lambda$. Actually, it dominates the first term at around
$\lambda=\tau_n$ in Fig.\ref{fig:verification} (a). The entire DOF of
SST is relatively large for small value of $\lambda$.
This implies that
the DOOF is serious for SST compared to ST when the threshold
level is small. The important point is that this effect is mainly
brought about by the zero true coefficients under a sparse setting in a
nonparametric regression. Therefore, the over-fitting behavior of SST is
notably different from that of ST around $\lambda=\tau_n$. From the
another point of view, ST suppresses this effect by an amount of
shrinkage even when the selection via thresholding is the same as the
manner of HT.

Apparently, we may say that the second term of DOF in SST encodes the
magnitudes of true coefficient values in adjusting $\lambda$. This is
because the slope of variation of the coefficient estimator is large at
around $\lambda$ as the Stein's lemma says and the probability measure
concentrates on around a true coefficient value. Actually, in
Fig.\ref{fig:verification} (a), we can see large values of the second
term at around $\lambda=0$ and also $\lambda=1$.  The former is due to 
zero true coefficient values and the latter is due to non-zero true
coefficient values; i.e. $b_k=1$ for all $k\in K^*$.

\subsection{Discussion on risk}

In Fig.\ref{fig:verification} (b), firstly, we can see that SURE is
almost consistent with the actual risk for ST and SST. Moreover, the
risk and SURE of SST approximate the risk of HT well. These two facts
support our theoretical result. We next consider the difference between
ST and SST.

As well known, HT and the bridge methods can reduce the bias that arises
in ST. This bias problem occurs because both of the threshold level and
the amount of shrinkage are simultaneously controlled by a single
parameter.  The bias harms the estimates of non-zero true components and
causes a high risk. Although this may interfere with the consistency of
thresholding; e.g. see \cite{HZ2006,ZY2006}, it is free for our
orthogonal setting. However, it is still a problem in the practical
model selection based on the risk that is a prediction error.  As found
in Fig.\ref{fig:verification} (b), the risks of HT and SST are minimized
at around $\lambda=0.5$ while the risk of ST is large at around this
value.  This is because of the bias mentioned above.  The important point
is that, in ST, a threshold level that minimizes the risk tends to be
small due to the bias problem.

On the other hand, for small threshold values, the risks of all methods
are high.  As easily understood, this is because the over-fitting to
noise.  At a very small threshold levels, risks are very high for all
methods due to the over-fitting originated from the number of unremoved
coefficients.  The important point is that the risks of HT and SST is
higher than that of ST. The difference is notable at between $0.05$ and
$0.1$, which is around $\tau_n$.  This fact corresponds to a high DOOF
due to a HT property as was found in Fig.\ref{fig:verification} (a).  We
can regard $\tau_n$ as a noise level for the coefficient estimate in some
sense. Therefore, we can say that the risks of HT and SST are high at
around the noise level. Although this is caused by over-fitting, it may
be preferable in model selection based on a risk estimate because a
high risk at around a noise level avoids the choice of a excessively
larger model that is in over-fitting.

As a result, SURE tends to choose a larger model for ST and a
smaller model for HT and SST. Note that, for a sparse and non-parametric
setting, a penalty for over-fitting is very high at around noise level in HT
and SST. Therefore, those may appropriately remove noise components in a
prediction error based model selection. 

\subsection{Model selection properties}

We here compare HT, ST, FT and SST in terms of risk,
sparseness and consistency of selection. We generate samples according
to (\ref{eq:y}) under the condition below. We set $n=256$ and
$\sigma^2=1$.  We consider the two case of a true representation.
\begin{itemize}
\setlength{\leftskip}{8mm}
 \item[Case-1] $K^*={1,2,3,4,5}$ and $b_k=1$ for $k\in K^*$.
 \item[Case-2] $K^*={1,2,\cdots,64}$ and $b_k=5/k$ for $k\in K^*$.
\end{itemize}
The first case assumes that a true representation is very sparse and it
can be easily identified. The second case assumes that a true
representation is moderately sparse while some components are difficult
to be identified. For each case, we estimate the coefficients by using
HT, ST, FT and SST under a fixed threshold level.  For HT, we employ the
universal thresholding level\cite{DJ1994,DJ1995}. We need the estimates
of noise variance for computing SURE and threshold level of HT.  It is
calculated by an unbiased estimate using the first $n/2$ components. It
is possible because of the setting of $K^*$.  However, we note that it
is difficult to obtain an unbiased estimate in general situations while
a suitable solution is give in wavelet
denoising\cite{DJ1994,DJ1995}. The candidates of the threshold level are
$\lambda\in\{0.02,0.03,\dots,0.1,0.2,\ldots,1\}$. FT and SST have
additional hyper-parameters $\gamma$ and $m$ respectively.  The
candidate of these parameters are $\gamma\in\{1.1,1.2,1.5,2,3,4,5\}$ and
$m\in\{1,3,5,7,9,11\}$ respectively. We choose a threshold level
according to SURE for HT and ST.  For FT and SST, we choose both of
threshold level and hyper-parameter by a grid search of SURE. Since we
know the true representation, we can calculate an actual risk. For
comparing model selection properties, we obtain the number of non-zero
coefficient estimates. Note that it is a measure of sparseness.  Also,
we measure a selection error by the cardinality of the symmetric
difference between $K^*$ and $\eK_{\lambda}$. We refer to these as Risk,
$\ek$ and SErr here. We repeat $S=5000$ trials and calculate the
averages of these values.  In Table. \ref{tbl:1}, we show the result, in
which we also append the standard deviation in the bracket.  We
summarize the results.

\begin{table}[b]
\caption{The averages of risk, $\ek$ and SErr.}
\label{tbl:1}
\begin{center}

 (a) Case-1
\begin{center} 
   \begin{tabular}{|c|c|c|c|}\hline
    & Risk & $\ek$ & SErr\\\hline
HT&0.0300(0.0257)&5.2090(0.4707)&0.2090(0.4707)\\\hline
ST&0.1164(0.0406)&39.2140(12.6461)&34.2140(12.6461)\\\hline
FT&0.0606(0.1199)&9.2940(10.3690)&4.2940(10.3690)\\\hline
SST&0.0384(0.0651)&7.1930(7.1939)&2.1930(7.1939)\\\hline
  \end{tabular}
\end{center}
\end{center}
\begin{center}
 (b) Case-2
\begin{center}  
\begin{tabular}{|c|c|c|c|}\hline
    & Risk & $\ek$ & SErr\\\hline
HT&0.7876(0.1058)&26.3850(3.1456)&37.9290(3.0498)\\\hline
ST&0.5393(0.0690)&126.7680(20.1387)&77.7920(15.8806)\\\hline
FT&0.6306(0.1529)&84.2580(20.7942)&47.0240(13.5550)\\\hline
SST&0.5600(0.1170)&81.0050(17.8764)&44.6370(11.4926)\\\hline
\end{tabular}
\end{center}
 \end{center}
\end{table}

\begin{itemize}
 \item For Case-1, HT is superior to the other methods in
       terms of all of risk, sparseness and consistency.
On the other hand, ST shows the worst performance that is caused by a
       large bias for contributed coefficient estimates. This induces a
       choice of a larger size as suggested by 
       Fig.\ref{fig:verification}.
SST and FT are inbetweens. Since a true representation is clearly
       identified in Case-1, it is easy to separate it from noise. The
       success of HT comes from this fact.

 \item For Case-2, ST gives the lowest risk while it does not give us a
       sparse and correct representation.  On the other hand, HT shows
       the worst risk value while it gives a highly sparse
       representation. This comes from the fact that there are many true
       non-zero coefficients that are hard to be identified due to their
       magnitudes. Of course, the hardness may depend on a signal to noise
       ratio and the number of samples.
HT tends to remove these weak components. SErr is relatively
       small for HT since HT may surely select the true components. We
       can see that ST may select redundant components while it gives a
       low risk. This is because many true components are included in
       the selected components. Again, SST and FT are inbetweens.

\end{itemize}
By these results, SST and FT may be reliable in applications since we do
not know the situation of a true representation.  The performances of SST
and FT are comparable. A certain advantage of SST is found in the
tables. This may come from the fact that FT connects ST to HT by a
non-smooth manner while SST does by a smooth manner.

\section{Conclusions and future works}

In this article, we developed and analyzed SST method in which ST
estimators are independently expanded by empirical scaling values.  The
scaling values have a common hyper-parameter that is an order of Taylor
expansion of an ideal scaling value that achieves HT.  The SST estimator
expands the ST estimator while it shrinks the least squares estimator.
SST estimator is a bridge estimator between ST and HT estimators and is
a generalized estimator that includes ST and NNG estimators as special
cases.  It also gives an another derivation of the well-known AL under a
specific regularization parameter and, therefore, it gives an
interpretation of AL.

We then derived the DOF of SST estimator by means of the SURE and found
that it is decomposed into the DOF of ST and the reminder connecting to
HT.  In this meaning, SST method gives a natural bridge between ST and
HT.  Since the DOF represents the DOOF, this result implies that there
are two sources of over-fitting in SST method. The first source
originated from ST is determined by the number of un-removed
coefficients and is a natural measure of the DOOF.  Since we showed that
the DOF of SST converges to that of HT as the expansion order goes to
infinity, we attempt to analyze the second source of HT as a limit case.
To do this, we showed a simple numerical example and explained the
numerical result based on the theoretical result.  In this example, we
showed the change of DOF, risk and SURE in terms of threshold levels.
We here could see the non-monotonicity of the second source of
over-fitting in DOF as pointed out in \cite{RT2014}. We then found the
non-monotonicity at a relatively small threshold level comes from a
over-fitting to noise.  More precisely, in a sparse and non-parametric
setting, the second source of over-fitting is largely determined by
coefficient estimates whose true values are zeros.  And, the impact is
maximized when a threshold level is at around noise levels in these
coefficient estimates.  This excess over-fitting given by the second
source leads to a large penalty in risk. Therefore, for SST, a larger
model that is in over-fitting tends to be excluded in a model selection
based on a risk estimate. On the other hand, as well known, for a large
threshold level, risk of ST is high due to a bias problem and SST are
free from this. As a result, SURE for ST tends to choose a smaller
threshold level that yields a larger model while SURE for SST tends to
choose a larger threshold level that yields a smaller model.  In an
actual model selection experiment, we showed that advantage of
thresholding method depends on the signal-to-noise ratio and the bridge
method such as SST and FT may be preferable compared to ST and HT in
general situations.

On the other hand, in thresholding methods, coefficients with large
contribution to fitting are not removed or, more actively, selected to
reduce the fitting error in greedy manner. Obviously, this can be a
source of over-fitting and it may be independent of model size; i.e. the
number of coefficients.  Indeed, a detailed asymptotic analysis in this
direction is found in \cite{KH2016}, in which the over-fitting mechanism
in a greedy method is shown to be related to the extreme value property
of the coefficient estimates of zero true values.  This is partly
consistent with our result here.  As a future work, we may need more
detailed correspondence between the result here and the results in
\cite{KH2016} that focus on the selectability of variables in the
fitting procedure. This point may be a part of the answer for an
exploration in \cite{RT2014}.

\section*{Acknowledgment}

This work was supported by Japan Society for the Promotion
of Science (JSPS) KAKENHI Grant Number 18K11433.


\end{document}